\documentclass[a4paper,11pt,fleqn]{article}
\pdfoutput=1
\usepackage{fullpage}
\usepackage{ucs}
\usepackage[utf8]{inputenc}
\usepackage{amsmath}
\usepackage{amsfonts}
\usepackage{amssymb}
\usepackage[english]{babel}
\usepackage[pdftex]{graphicx}

\usepackage[pdftex]{hyperref}

\title{Deep Information Networks}
\author{Giulio Franzese, Monica Visintin\\
 \\Dipartimento di Elettronica e Telecomunicazioni\\ Politecnico di Torino, Italy\\
\texttt{giulio.franzese{@}polito.it}, \texttt{monica.visintin{@}polito.it}
}
\date{}

\begin{document}
\maketitle
\begin{abstract}
We describe a novel classifier with a tree structure, designed using information theory concepts. This Information Network  is made of information nodes, that compress the input data, and multiplexers, that connect two or more input nodes to an output node. Each information node is trained, independently of the others, to minimize a local cost function that minimizes the mutual information between its input and output with the constraint of keeping a given mutual information between its output and the target (information bottleneck). We show that the system is able to provide good results in terms of accuracy, while it shows many advantages in terms of modularity and reduced complexity.
\end{abstract}


\section{Introduction}
\label{sect:intro}
The so-called ``information bottleneck'' was described in \cite{tishby,tishby2}, where it was employed to analyze how the information flows inside a classification  neural network. We here propose to exploit the same idea to build a supervised learning machine that compresses the input data and generates the estimated class, using a modular structure with a tree topology. 

The key element of the proposed Deep Information Network (DIN) is the information node, that compresses the input while keeping its information content: during the training phase the node evaluates the conditional probabilities $P(X_{out}=j|X_{in}=i)$ that minimize the mutual information $\mathbb{I}(X_{in};X_{out})$ between its input $X_{in}$ and output $X_{out}$ for a given mutual information between its output and the target $\mathbb{I}(X_{in};Y)$. The input of each information node at the first layer is one of the features/attributes of the data; once the conditional probabilities are found, the node randomly generates its output according to them. 

The second element of the DIN is the
multiplexer that combines the outputs of two or more information nodes to generate the input of the subsequent one. The machine is therefore made of several layers, and the number of information nodes per layer decreases from layer to layer, thanks to the presence of multiplexers, until a unique information node is obtained in the last layer, whose task is to provide the estimated class of the input. 

Thus a DIN is similar to a neural network, but the working principle is completely different. In particular, there is no global objective function and each information node is separately trained, using only its input and the target class. The layers are trained in order: the nodes of the first layer are trained using the available raw data (one node for each of the available features) and stochastically generate the data for the multiplexers that feed the second layer; the nodes at the second layer are trained using their input (ideally the information contained in two or more features in the raw data) and the target, etc. Each information node has no knowledge about the input and output of other nodes. From layer to layer, the information nodes and the multiplexers allow to extract the information content of the original data as if they were sifting it. Whereas algorithm C4.5 by Quinland \cite{Quinland} builds a hierarchical decision tree starting from the root, we here propose a way to build a similar tree, but starting from the leaves.

The advantages of DINs are clear: extreme flexibility, high modularity, complexity that linearly depends on the number of nodes, overall number of nodes that linearly depends on the number of features of the data. Since the machine works with information theory, categorical and missing data are easily managed; on the other hand, continuous data should be first quantized, but the true quantization can be left to the first layer information nodes. 

What has to be assessed for the proposed machine is the performance: does it provide good results in terms of misclassification probability? are there constraints on its parameters? 

Section \ref{sec:network} more precisely describes the structure of the DIN, section \ref{sec:analysis} gives some insight on the theoretical properties and section \ref{sec:results} comments the results obtained with a dataset available on the web. Conclusions are finally drawn in section \ref{sec:concl}.

\section{The information network}
\label{sec:network}

As pointed out in \ref{sect:intro}, the machine is made of information nodes, described in section \ref{sec:infonode}, and multiplexers joined together through a network described in section \ref{sec:netw_arch}.

\subsection{The information node}
\label{sec:infonode}
\begin{figure}[b]
\begin{center}
 \includegraphics[width=0.2 \textwidth]{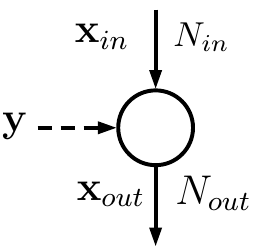}
\end{center}
\label{fig:infonode} 
\caption{Schematic representation of an information node, showing the input and output vectors, with alphabets of cardinality $N_{in}$ and $N_{out}$, respectively, and the target vector $\mathbf{y}$ 
which is available during the training phase.}
\end{figure}

Each information node, described in Fig. \ref{fig:infonode}, has an input vector $\mathbf{x}_{in}$, whose elements take values in a set of cardinality $N_{in}$, and an output vector $\mathbf{x}_{out}$, whose elements take values in a set of cardinality $N_{out}$; moreover the target class vector $\mathbf{y}_{train}$ is available as input, with elements in the set $[0,N_{class}-1]$; all the vectors have the same size $N_{train}$. 
By setting $N_{out}<N_{in}$, the information node performs a compression (source encoding). 
Let $X_{in}$ ($X_{out}$, $Y$) identify the random variable whose samples are stored in vector $\mathbf{x}_{in}$ ($\mathbf{x}_{out}$, $\mathbf{y}$). 

In the training phase, the information node randomly generates the output vector $\mathbf{x}_{out}$, according to the conditional probability mass function that satisfies equation  \cite{tishby}
\begin{equation}
P(X_{out}=j|X_{in}=i)=\frac{1}{Z(i;\beta)}P(X_{out}=j)e^{-\beta d(i,j)}, \quad i=0,\ldots,N_{in}-1, j=0,\ldots,N_{out}-1
\label{eq:constraint}
\end{equation}
where 
\begin{itemize}
\item $P(X_{out}=j)$ is the probability mass function of the output random variable $X_{out}$
\begin{equation}
P(X_{out}=j)=\sum_{i=0}^{N_{in}-1}P(X_{in}=i)P(X_{out}=j|X_{in}=i),\quad  j=0,\ldots,N_{out}-1
\end{equation}
\item $d(i,j)$ is the Kullback-Leibler divergence
\begin{align}
d(i,j)&=\sum_{m=0}^{N_{class}-1} P(Y=m|X_{in}=i)\log_2 \frac{P(Y=m|X_{in}=i)}{P(Y=m|X_{out}=j)}\nonumber \\
&=\mathbb{KL}(P(Y|X_{in}=i)||P(Y|X_{out}=j))
\end{align}
and
\begin{align}
P(Y=m|X_{out}=j)=\sum_{i=0}^{N_{in}-1}&P(Y=m|X_{in}=i)P(X_{in}=i|X_{out}=j),\nonumber \\
& m=0,\ldots,N_{class}-1, j=0,\ldots,N_{out}-1
\end{align}

\item $\beta$ is a real positive parameter
\item $Z(i;\beta)$ is a normalizing coefficient that allows to get
\begin{equation}
\sum_{j=1}^{N_{out}-1}P(X_{out}=j|X_{in}=i)=1
\end{equation}
\end{itemize}
The probabilities $P(X_{out}=j|X_{in}=i)$ can be iteratively found using the Blahut-Arimoto algorithm \cite{tishby}. 

According to \cite{tishby}, Eqn. (\ref{eq:constraint}) solves the information bottleneck: it minimizes the mutual information $\mathbb{I}(X_{in};X_{out})$ under the constraint of a given mutual information $\mathbb{I}(Y;X_{out})$. In particular, eqn. (\ref{eq:constraint}) is the solution of the minimization of the Lagrangian
\begin{equation}
\mathcal{L}=\mathbb{I}(X_{in};X_{out})-\beta \mathbb{I}(Y;X_{out})
\end{equation}
If the Lagrangian multiplier $\beta$ is increased, then the constraint is privileged and the information node tends to maximize the mutual information between its output $X_{out}$ and the class $Y$; if $\beta$ is reduced, then minimization of $\mathbb{I}(X_{in};X_{out})$ is obtained, thus a compression is obtained. The information node must actually balance compression from $X_{in}$ to $X_{out}$ and propagation of the information on $Y$, in order to correctly decide its value. In our implementation, the compression is also imposed by the fact that the cardinality of the output alphabet $N_{out}$ is smaller than that of the input alphabet $N_{in}$.

\subsection{The network}
\label{sec:netw_arch}
\begin{figure}
 \begin{center}
 \includegraphics[width=0.9 \textwidth]{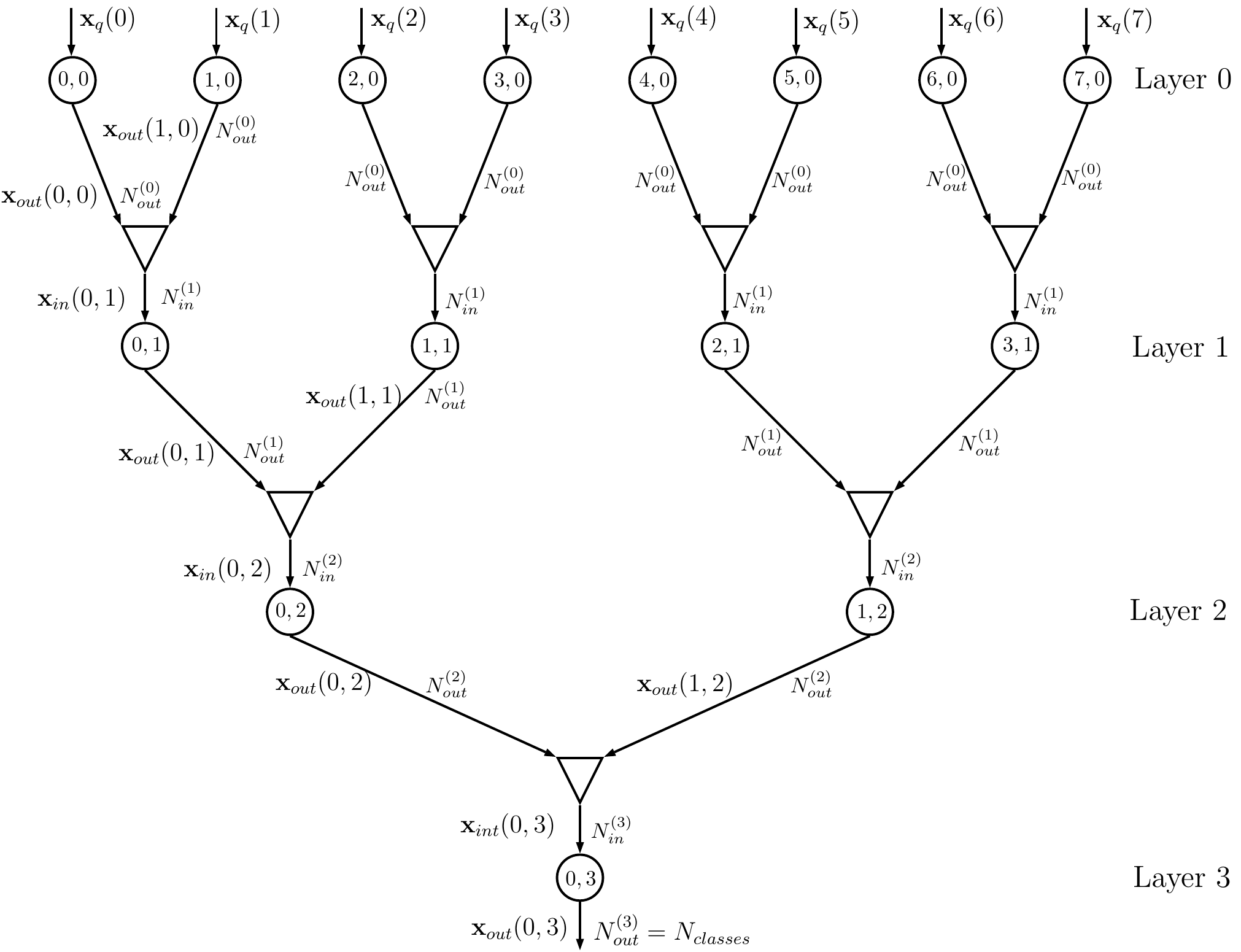}
 % schematic.pdf: 0x0 pixel, 300dpi, 0.00x0.00 cm, bb=
\end{center}
\caption{Architecture of a simple information node network for $D=8$: each info node is represented by a circle, the numbers inside the circle identify the node, the triangles identify the mixers, $N_{in}^{(k)}$ is the number of values taken by the input of the info node at layer $k$, $N_{out}^{(k)}$ is the number of values taken by the output of the info node at layer $k$.}
\label{fig:arch1}
\end{figure}

\begin{figure}
\begin{center}
 \includegraphics[height=0.3 \textheight,angle=90]{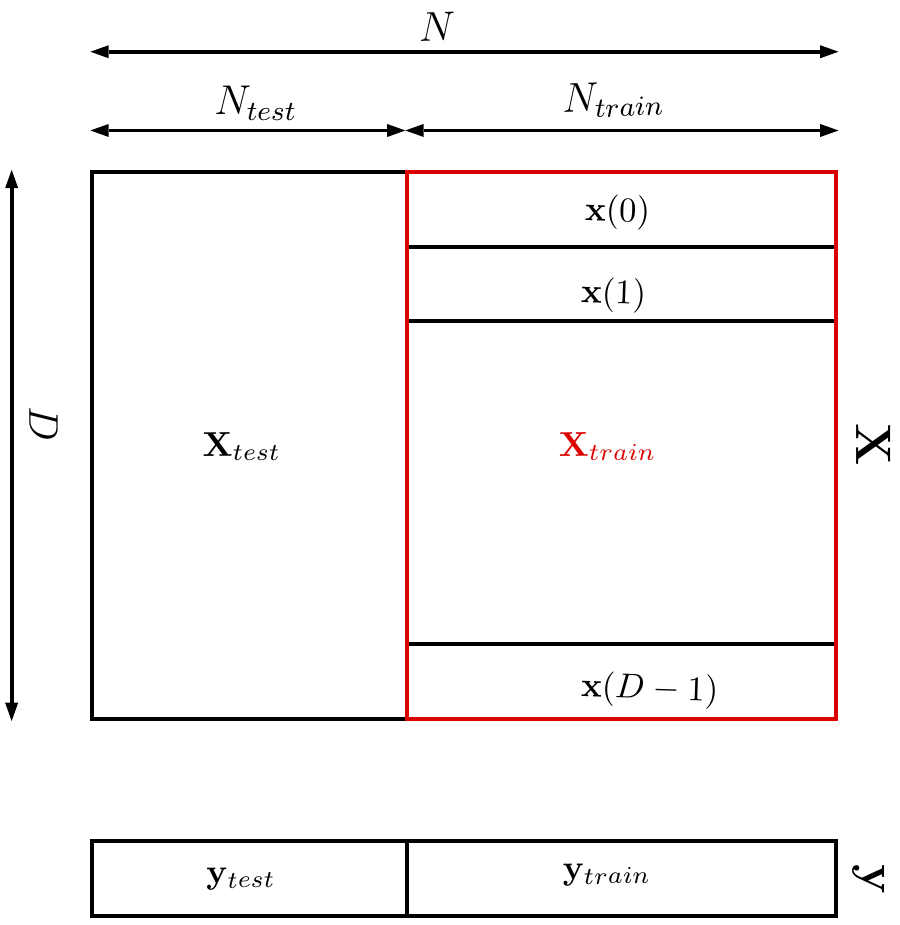}
\end{center}
\caption{Definition of matrices, vectors and their dimensions}
\label{fig:matrvect}
\end{figure}

Fig. \ref{fig:arch1} shows an example of a DIN, where
we assume that the dataset is made of a matrix $\mathbf{X}$ with $N$ rows and $D=8$ columns (features) and the corresponding class vector $\mathbf{y}$; we use $N_{train}$ rows for the training phase and $N_{test}=N-N_{train}$ rows for the testing phase, getting matrices $\mathbf{X}_{train}$ and $\mathbf{X}_{test}$ (see Fig. \ref{fig:matrvect}). 

Now refer to Fig. \ref{fig:arch1}.
The $k$-th column $\mathbf{x}(k)$ of matrix $\mathbf{X}_{train}$ should be, together with vector $\mathbf{y}_{train}$, the input of the information node $(k,0)$ at layer 0 of the structure. However the algorithm needs a finite number $N_{in}^{(0)}$ of values in the input vectors, and a pre-processing quantization phase is needed. Quantization (uniform, Max-Lloyd, etc.) is not discussed here, in the example described in Sect \ref{sec:results} we used linear quantization. Thus the true input of node $(k,0)$ is $\mathbf{x}_q(k)$, the quantized version of  $\mathbf{x}(k)$. Of course quantization is not necessary for categorical data. The number of nodes at layer 0 of the structure is $D$, equal to the number of columns of $\mathbf{X}$.

Information node $(k,0)$ at layer 0 thus processes  $\mathbf{x}_q(k)$ and $\mathbf{y}_{train}$, and generates the output vector $\mathbf{x}_{out}(k,0)$ with alphabet of cardinality $N_{out}^{(0)}$. Note that this output vector is randomly generated by the information node, according to its probability matrix $\mathbf{P}_{k,0}$, whose element $i,j$ is $P(X_{out}=j|X_{in}=i)$ as found by the algorithm described in section \ref{sec:infonode}.

The output vectors $\mathbf{x}_{out}(2k,0)$ and $\mathbf{x}_{out}(2k+1,0)$ are combined together by a multiplexer (shown as a triangle in Fig. \ref{fig:arch1}) that outputs
\begin{equation} 
\mathbf{x}_{in}(k,1)=\mathbf{x}_{out}(2k,0)\;+\;N_{out}^{(0)}\;\mathbf{x}_{out}(2k+1,0)\end{equation}
Vector $\mathbf{x}_{in}(k,1)$ is now the input of the information node $(k,1)$ of layer 1, and has an alphabet with cardinality
\begin{equation} 
N_{in}^{(1)}=N_{out}^{(0)}\;\times \;N_{out}^{(0)} 
\end{equation}

The topology is iterated halving the number of information nodes from layer to layer, until an isolated information node is obtained at layer $d=\log_2(D)$, that has input vector $\mathbf{x}_{in}(0,d)$ with alphabet cardinality $N_{in}^{(d)}$ and output vector $\mathbf{x}_{out}(0,d)$, whose alphabet cardinality must be $N_{out}^{(d)}=N_{class}$, the number of classes in the target vector $\mathbf{y}_{train}$. If the parameters are correctly chosen, the output vector $\mathbf{x}_{out}(0,d)$ is equal to $\mathbf{y}_{train}$, and the probability matrix $\mathbf{P}_{d,0}$ has just one value equal to 1 in each of its rows. A tree topology is thus obtained.

In the overall, the topology proposed in Fig. \ref{fig:arch1} requires a number of information nodes equal to
\begin{equation}
N_{nodes}= D+\frac{D}{2}+\frac{D}{4}+\cdots+2+1=2D-1
\end{equation}
and a number of mixers equal to
\begin{equation}
N_{mixers}=\frac{D}{2}+\frac{D}{4}+\cdots+2+1=D-1
\end{equation}
All the nodes run exactly the same algorithm and all the mixers are equal, apart from the input/output vector alphabet cardinalities. If the cardinalities of the alphabets are all equal, i.e. $N_{in}^{(i)}$ and $N_{out}^{(i)}$ do not depend on the layer $i$, then all the nodes and all the mixers are exactly equal, which might help in a possible hardware implementation.

\subsection{The testing/running phase}
During the testing or running phase, the quantized columns of matrix $\mathbf{X}_{test}$ are used as inputs and each node simply generates the output vector $\mathbf{x}_{out}$ according to the input vector $\mathbf{x}_{in}$, using the probability matrix $\mathbf{P}$ that has been optimized during the training phase. For example, if the $n$-th value stored in the vector $\mathbf{x}_{in}(k,\ell)$ of node $k,\ell$ is $i$, then the $n$-th value in vector $\mathbf{x}_{out}(k,\ell)$ is $j$ with probability $\mathbf{P}_{k,\ell}(i,j)$. Each node then generates a stochastic output, according to its probability matrix.

\section{Analysis}
\label{sec:analysis}
In the following, the DIN is studied considering first the overall conditional probabilities between the input features and the output, and second the information flow inside it.
\subsection{The probabilistic point of view}

\begin{figure}
\begin{center}
 \includegraphics[height=0.3 \textheight]{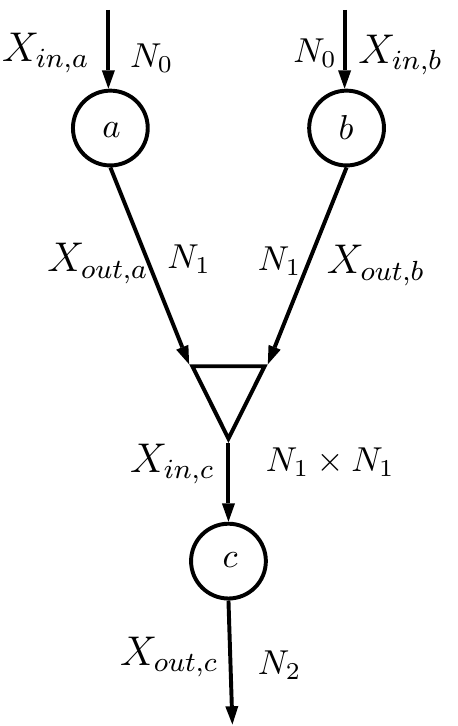}
\end{center} 
\caption{Sub-network used for the evaluation of the probability matrix; $X_{in,a}$, $X_{out,a}$, $X_{in,b}$, $X_{out,b}$, $X_{in,c}$, $X_{out,c}$ are all random variables; $N_0$ is the number of values taken by $X_{in,a}$ and $X_{in,b}$, $N_1$ is the number of values taken by $X_{out,a}$ and $X_{out,b}$, $N_2$ is the number of values taken by $X_{out,c}$.}
\label{fig:subnet}
\end{figure}

Consider the case of the sub-network made of nodes $a$, $b$, $c$ as shown in Fig. \ref{fig:subnet}; the alphabet cardinality at the input of nodes $a$ and $b$ is $N_0$, the cardinality at the output of nodes $a$ and $b$ is $N_1$, the alphabet cardinality at the input of node $c$ is $N_1 \times N_1$, the cardinality at its output is $N_2$. Node $a$ is characterized by matrix $\mathbf{P}_a$, whose element $P_a(i,j)$ is $P(X_{out,a}=j|X_{in,a}=i)$; similar definitions hold for   $\mathbf{P}_b$ and $\mathbf{P}_c$. Note that $\mathbf{P}_a$ and $\mathbf{P}_b$ have $N_0$ rows and $N_1$ columns, whereas 
$\mathbf{P}_c$ has $N_1\times N_1$ rows and $N_2$ columns; the overall probability matrix between the inputs $X_{in,a}$, $X_{in,b}$ and the output $X_{out,c}$ is $\mathbf{P}$ with $N_0\times N_0$ rows and $N_2$ columns.
Then
\begin{small}\begin{align}
&P(X_{out,c}=i|X_{in,a}=j,X_{in,b}=k) \nonumber\\
&= \sum_{r=0}^{N_1-1} \sum_{s=0}^{N_1-1}  P(X_{out,c}=i,X_{out,a}=r,X_{out,b}=s|X_{in,a}=j,X_{in,b}=k)\nonumber\\
&=\sum_{r=0}^{N_1-1} \sum_{s=0}^{N_1-1} P(X_{out,c}=i|X_{out,b}=r,x_{out,c}=s)P(X_{out,a}=r|X_{in,a}=j)P(X_{out,b}=s|X_{in,b}=k)\nonumber\\
&=\sum_{r=0}^{N_1-1} \sum_{s=0}^{N_1-1} P(X_{out,c}=i|X_{out,b}=r,X_{out,c}=s)\mathbf{P}_a(j,r)\mathbf{P}_b(k,s)
\end{align}
\end{small}% 
It can be shown that
\begin{equation}
\mathbf{P}=(\mathbf{P}_a \otimes \mathbf{P}_b)\mathbf{P}_c
\end{equation}
where $\otimes$ identifies the Kronecker matrix multiplication; note that $\mathbf{P}_a \otimes \mathbf{P}_b$ has $N_0 \times N_0$ rows and $N_1 \times N_1$ columns.
By iteratively applying the above rule, we can get the expression of the overall matrix $\mathbf{P}$ for the exact topology of Fig. \ref{fig:arch1}, with 8 input nodes and four layers:
\begin{align}
\mathbf{P}&=\bigg[\Big\{ \big[(\mathbf{P}_{0,0} \otimes \mathbf{P}_{1,0})\mathbf{P}_{0,1}\big]\otimes\big[(\mathbf{P}_{2,0} \otimes \mathbf{P}_{3,0})\mathbf{P}_{1,1}\big]\Big\}\mathbf{P}_{0,2}  \nonumber \\
&\otimes  \Big\{ \big[(\mathbf{P}_{4,0} \otimes \mathbf{P}_{5,0})\mathbf{P}_{2,1}\big]\otimes\big[(\mathbf{P}_{6,0} \otimes \mathbf{P}_{7,0})\mathbf{P}_{3,1}\big]\Big\}\mathbf{P}_{1,2} \bigg]\mathbf{P}_{0,3}
\end{align}
The DIN then behaves like a one-layer system that generates the output according to matrix $\mathbf{P}$, whose size might be impractically large (with $D$ features all quantized with $N_0$ levels, matrix $\mathbf{P}$ has size $N_0^D \times 2$). The proposed layered structure needs smaller probability matrices, which makes the system computationally efficient.
\subsection{The information theory point of view}

Each information node compresses the data and therefore, by a basic application of data processing inequality, we have
\begin{equation}
\mathbb{I}(X_{out};Y) \ge \mathbb{I}(X_{in};Y)
\end{equation}

Consider now the outputs $X_{out}(2k,i)$ and $X_{out}(2k+1,i)$ of two nodes at layer $i$ that are combined by the mixer to generate $X_{in}(k,i+1)$. It is interesting for analysis and integrity purposes to derive an upper and a lower bound of the total amount of information obtained at the output of the mixer. First of all notice that, from an information theoretic point of view, we can write 
\begin{equation}
\label{eq:ip1}
\mathbb{I}(X_{in}(k,i+1);Y)=\mathbb{I}([X_{out}(2k,i),X_{out}(2k+1,i)];Y)
\end{equation}
We can expand \eqref{eq:ip1} as
\begin{align}
\mathbb{I}([X_{out}(2k,i)&,X_{out}(2k+1,i)];Y)\nonumber\\
&=\mathbb{H}([X_{out}(2k,i),X_{out}(2k+1,i)])-\mathbb{H}([X_{out}(2k,i),X_{out}(2k+1,i)]|Y)\nonumber\\
&=\mathbb{H}(X_{out}(2k+1,i)|X_{out}(2k,i))+\mathbb{H}(X_{out}(2k,i))\nonumber\\
&-(\mathbb{H}(X_{out}(2k+1,i)|X_{out}(2k,i),Y)-\mathbb{H}(X_{out}(2k,i))|Y)\nonumber\\
&=\mathbb{I}(X_{out}(2k,i);Y)+\mathbb{I}(X_{out}(2k+1,i)|X_{out}(2k,i);Y)\label{eq:ip2}\\
&=\mathbb{I}(X_{out}(2k+1,i);Y)+\mathbb{I}(X_{out}(2k,i)|X_{out}(2k+1,i);Y)\label{eq:ip3}
\end{align}
where the last equality is due to evident symmetries in the equations. Combining \eqref{eq:ip2} and \eqref{eq:ip3}, and due to non-negativity of mutual information, we obtain
\begin{align}
\label{eq:ip4}
\mathbb{I}(X_{in}(k,i+1);Y)& \ge \max\{\mathbb{I}(X_{out}(2k,i);Y), \mathbb{I}(X_{out}(2k+1,i);Y) \}
\end{align}
where equality is attained when $\mathbb{H}(X_{out}(2k+1,i)|X_{out}(2k,i))=0$ (i.e. when $X_{out}(2k+1,i)$ can be exactly predicted if $X_{out}(2k,i)$ is known).\\
We can also derive an upper bound for the total mutual information. First we observe that
\begin{align}
&\mathbb{H}([X_{out}(2k,i),X_{out}(2k+1,i)])\leq \mathbb{H}(X_{out}(2k,i))+\mathbb{H}(X_{out}(2k+1,i))\label{eq:ip5}\\
&\mathbb{H}([X_{out}(2k,i),X_{out}(2k+1,i)]|Y)\geq \max\{\mathbb{H}(X_{out}(2k,i)|Y),\mathbb{H}(X_{out}(2k+1,i)|Y) \}\label{eq:ip6}
\end{align}
and thus we can derive, using \eqref{eq:ip5} and \eqref{eq:ip6}, that
\begin{align}
\mathbb{I}(X_{in}(k,i+1);Y)&=
\mathbb{H}([X_{out}(2k,i),X_{out}(2k+1,i)])-\mathbb{H}([X_{out}(2k,i),X_{out}(2k+1,i)]|Y)\nonumber\\
&\leq \mathbb{H}([X_{out}(2k,i),X_{out}(2k+1,i)])-\mathbb{H}(X_{out}(2k,i)|Y)\nonumber\\
&\leq \mathbb{H}(X_{out}(2k,i))+\mathbb{H}(X_{out}(2k+1,i))-\mathbb{H}(X_{out}(2k,i)|Y)\nonumber\\
&=\mathbb{I}(X_{out}(2k,i);Y)+\mathbb{H}(X_{out}(2k+1,i))\label{eq:ip7}\\
&=\mathbb{I}(X_{out}(2k+1,i);Y)+\mathbb{H}(X_{out}(2k,i))\label{eq:ip8}
\end{align}
where again the last equality is due to symmetry. We can then combine \eqref{eq:ip7} and \eqref{eq:ip8} in
\begin{align}
&\mathbb{I}(X_{in}(k,i+1);Y)\leq \nonumber\\
&\min\{\mathbb{I}(X_{out}(2k,i);Y)+\mathbb{H}(X_{out}(2k+1,i)),\mathbb{I}(X_{out}(2k+1,i);Y)+\mathbb{H}(X_{out}(2k,i)) \}\label{eq:ip9}
\end{align}
Finally, combining \eqref{eq:ip4} and \eqref{eq:ip9}
\begin{align}
& \max\{\mathbb{I}(X_{out}(2k,i);Y), \mathbb{I}(X_{out}(2k+1,i);Y) \} \leq \nonumber\\
&\mathbb{I}(X_{in}(k,i+1);Y)\leq \nonumber\\
&\min\{\mathbb{I}(X_{out}(2k,i);Y)+\mathbb{H}(X_{out}(2k+1,i)),\mathbb{I}(X_{out}(2k+1,i);Y)+\mathbb{H}(X_{out}(2k,i)) \}\label{eq:ip10}
\end{align}
obtaining an upper and lower bound for the mutual information.

\section{The Kidney Disease experiment}
\label{sec:results}
In this section we present the results of an experiment performed on the UCI Kidney Disease dataset \cite{Lichman:2013,Salekin}. The dataset has a total of 24 medical features, consisting of mixed categorical, integer and real values, with missing values present. The aim of the experiment is to correctly classify patients affected by chronic kidney disease. Remember that the machine works also with missing values, since they are seen a a special categorical value, whose probability is the only required measure.
 
In the proposed information network, layer zero has as many information nodes as the number of features (i.e. 24), and the input of each node is one of the quantized features; these nodes are trained in parallel. Then the outputs of layer zero are mixed two at a time and they become the input of layer one with 12 information nodes. These 12 nodes are merged into 6 nodes into layer two and the 6 nodes are mixed into 3 nodes in layer three. Finally the three nodes are combined into the unique final node whose output cardinality is equal to 2 (corresponding to ill or healthy). The 24 input features are uniformly quantized with different values of $N_{in}^{(0)}$, depending on the feature (from 2 to 470); on the contrary the value of $N_{out}$ is 3 for all the nodes apart from the last one. The multiplexers generate vectors taking values in the range $[0,8]$, and the information nodes compress these values in the range $[0,2]$. The value of $\beta$ is equal to 5; the number of training points is equal to $N_{train}=200$ (half of which positive), the number of testing points is again $N_{test}=200$.

\begin{figure}
\begin{center}
 \includegraphics[height=7.5cm]{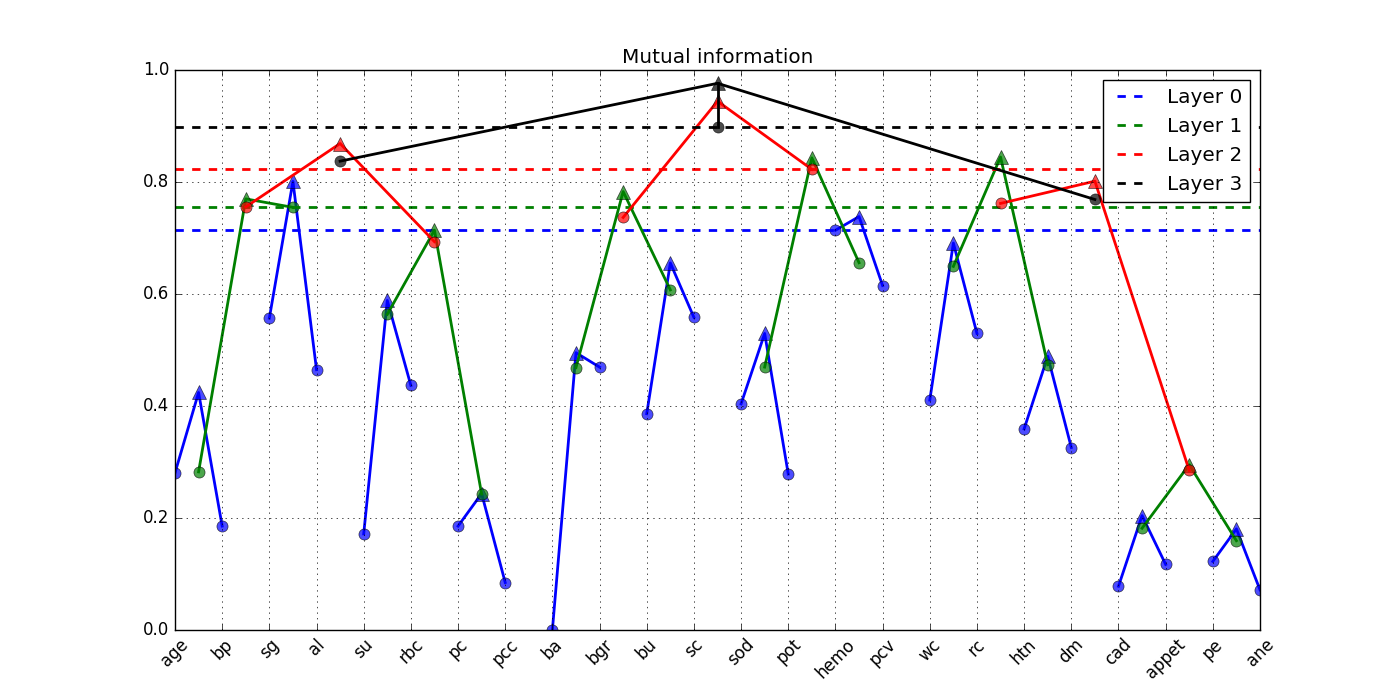}
\end{center}
\caption{Mutual information in the network: circles correspond to inputs of the multiplexers, triangles to their outputs, each dotted line represents the maximum value of the mutual information at the input of the multixers of a given layer. Results obtained during one run of the training phase with $N_{train}=200$ (half ill, half healthy), $N_{out}^{(i)}=3$.}
\label{fig:kidney1}
\end{figure}

Figure \ref{fig:kidney1} shows how the mutual information $\mathbb{I}(X;Y)$ evolves in the DIN, and it is possible to appreciate that it increases from the first to the last layer. In particular $\mathbb{I}(X_{out}(2k,i);Y)$ and $\mathbb{I}(X_{out}(2k+1,i);Y)$, represented as circles, are linked through segments to $\mathbb{I}(X_{in}(k,i+1);Y)$, represented as a triangle, to show that the lower bound in (\ref{eq:ip10}) holds from $i=0$ (blue markers), to $i=3$ (black markers). Figure \ref{fig:kidney1} also shows that $\mathbb{I}(X_{in}(k,1);Y)$ (blue triangle) is higher than $\mathbb{I}(X_{out}(k,1);Y)$ (green circle at the same x-value), which is due to the compression performed by the information node, that reduces the cardinality of $X_{in}(k,1)$ to just three values; this is true in general when $\mathbb{I}(X_{in}(k,i);Y)$ is compared to $\mathbb{I}(X_{out}(k,i);Y)$ for all values of $k$ and $i$.

\begin{table}
\begin{center}
\begin{tabular}{|c|c|c|c|c|c|c|c|}
\hline
& $N_{out}^{(i)}$ &  $N_{train}$ & $N_{test}$ & Accuracy &  Sensitivity &Specificity & F1 score\\
\hline
training & 3 & 200 & 200 & 0.9970 & 0.9948 & 0.9993 & 0.9970\\
testing & 3 & 200 & 200 & 0.9303 & 0.9188 & 0.9343 & 0.9260\\
\hline
training & 2 & 200 & 200 & 0.9924 & 0.9985 & 0.9963 & 0.9974\\
testing & 2 & 200 & 200 & 0.9648 & 0.9400 & 0.9734 & 0.9560\\
\hline
training & 3 & 320 & 80 & 0.9947 & 0.9868 & 0.9996 & 0.9932\\
testing & 3  & 320 & 80 & 0.9705 & 0.9435 & 0.9875 & 0.9647\\
\hline
training & 2 & 320 & 80 & 0.9924 & 0.9820 &0.9986  &0.9902 \\
testing & 2  & 320 & 80 & 0.9762 & 0.9511 & 0.9918 & 0.9709\\
\hline
\end{tabular}  
\end{center} 
\caption{Results obtained with DINs on the Kidney Disease dataset; averages over 1000 runs; $\beta=5$, the specified value of $N_{out}^{(i)}$ is valid for $i=0,1,2,3$, the last layer has $N_{out}^{(4)}=N_{classes}=2$.}
\label{ta:one}
\end{table} 

In terms of performance, the results listed in table \ref{ta:one} were obtained:
\begin{itemize}
 \item some form of overfitting is present when $N_{out}^{(i)}=3$ and better results are obtained when $N_{out}^{(i)}=2$, $i=1,\ldots, 3$
 \item accuracy, specificity and sensitivity are high in all the considered cases
 \item when $N_{out}^{(i)}=2$, the accuracy obtained with $N_{train}=200$ is similar to that obtained with $N_{test}=320$, which shows that 200 
 patients are enough to train the DIN
 \item results obtained with $N_{train}=320$ and $N_{test}=80$ can be comparable with those obtained in \cite{Salekin} using other algorithms (k-nearest neighbour, random forest, neural networks)
\end{itemize}
Notice that the DINs were run one thousand times, each time randomly changing the subsets of patients used for the training and the testing phases; table \ref{ta:one} shows the average results.

\section{Conclusions}
\label{sec:concl}
The proposed Deep Information Network (DIN) shows good results in terms of accuracy and represents a new simple modular structure, flexible and useful for various applications. Further optimization might be obtained by using a different value of $N_{out}$ for each of the information nodes (not necessarily all equal at each layer) and appropriately selecting the value of the Lagrangian multiplier $\beta$.


\begin{thebibliography}{9}
\bibitem{tishby} N.\,Tishby, N.\,Zaslavsky, \textit{Deep Learning and the Information Bottleneck Principle}  arXiv:1503.02406v1 [cs.LG],  March 2015
\bibitem{tishby2} N.\,Tishby, F.\,C.\,Pereira, W.\,Bialek  \textit{The information bottleneck method} arXiv:physics/0004057v1,  April 2000
\bibitem{Quinland} J.\,R.\,Quinlan, \textit{C4.5: Programs for Machine Learning}. Morgan Kaufmann Publishers, 1993
\bibitem{cover} M.\,Cover, J.\,A.\,Thomas, \textit{Elements of Information Theory}, 2nd Edition (Wiley Series in Telecommunications and Signal Processing), 2006
\bibitem{Lichman:2013} M. Lichman,  {UCI} Machine Learning Repository, University of California, Irvine, School of Information and Computer Sciences, 2013 \url{http://archive.ics.uci.edu/ml}
\bibitem{Salekin} A.\,Salekin, J.\,Stankovic, \textit{Detection of Chronic Kidney Disease and Selecting
Important Predictive Attributes},  2016 IEEE International Conference on Healthcare Informatics (ICHI),
4-7 Oct. 2016,  DOI: 10.1109/ICHI.2016.36 
\end{thebibliography}
\end{document}